\documentclass[sigconf,natbib=true]{acmart}
\AtBeginDocument{%
  }

\setcopyright{acmlicensed}
\copyrightyear{2018}
\acmYear{2018}
\acmDOI{XXXXXXX.XXXXXXX}
\acmConference[Conference acronym 'XX]{Make sure to enter the correct
  conference title from your rights confirmation email}{June 03--05,
  2018}{Woodstock, NY}
\acmISBN{978-1-4503-XXXX-X/2018/06}
\usepackage{enumitem}
\usepackage{tabularx}
\usepackage[table,xcdraw]{xcolor}
\usepackage{hyperref}

\begin{document}

\newcommand{\xyz}[1]{\textcolor{red}{#1}} 
\newcommand{\methodnameshort}{SteerX~}

\title{SteerX: Disentangled Steering for LLM Personalization}




\author{Xiaoyan Zhao\textsuperscript{1}, Ming Yan\textsuperscript{1}, Yilun Qiu\textsuperscript{1}, Haoting Ni}
\thanks{1 Equal contribution}
\author{Yang Zhang, Fuli Feng, Hong Cheng, Tat-Seng Chua}
\email{xzhao@se.cuhk.edu.hk, {ym689, nhting}@mail.ustc.edu.cn, {qiuyilun, zhangy}@nus.edu.sg}
\affiliation{%
  \institution{The Chinese University of Hong Kong, University of Science and Technology of China, National University of Singapore}
  \country{}
}

\renewcommand{\shortauthors}{Xiaoyan Zhao et al.}

\begin{abstract}
Large language models (LLMs) have shown remarkable success in recent years, enabling a wide range of applications, including intelligent assistants that support users' daily life and work. A critical factor in building such assistants is personalizing LLMs, as user preferences and needs vary widely. Activation steering, which directly leverages directions representing user preference in the LLM activation space to adjust its behavior, offers a cost-effective way to align the model's outputs with individual users. However, existing methods rely on all historical data to compute the steering vector, ignoring that not all content reflects true user preferences, which undermines the personalization signal. To address this, we propose \textbf{SteerX}, a disentangled steering method that isolates preference-driven components from preference-agnostic components. Grounded in causal inference theory, SteerX estimates token-level causal effects to identify preference-driven tokens, transforms these discrete signals into a coherent description, and then leverages them to steer personalized LLM generation. By focusing on the truly preference-driven information, SteerX produces more accurate activation steering vectors and enhances personalization. Experiments on two representative steering backbone methods across real-world datasets demonstrate that SteerX consistently enhances steering vector quality, offering a practical solution for more effective LLM personalization. 
\end{abstract}


\received{20 February 2007}
\received[revised]{12 March 2009}
\received[accepted]{5 June 2009}

\maketitle

\section{Introduction}

\begin{figure}[t]
    \centering
    \includegraphics[width=\linewidth]{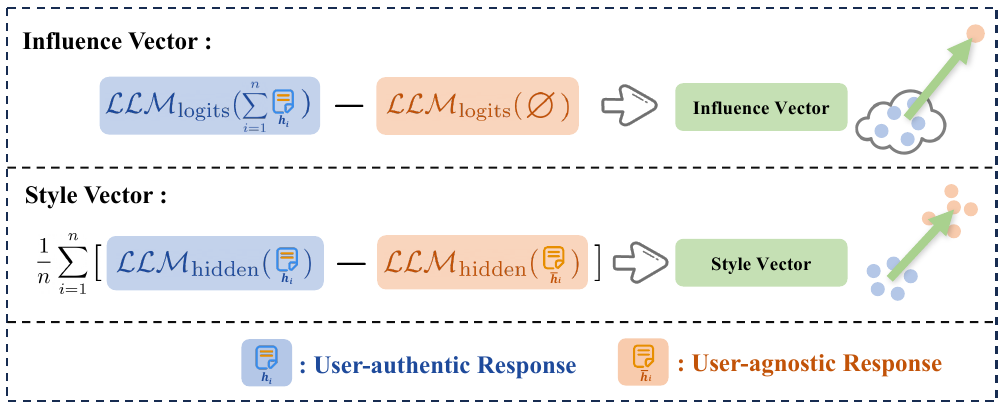}
    \caption{Two representative training-free activation steering methods. \emph{Influence Vector} is the logit difference with vs. without history. \emph{Style Vector} is the mean hidden-state difference between user-authentic response $h_i$ and user-agnostic response $\bar{h}_i$.}
    \label{fig:comparison}
    \vspace{-1.2em}
\end{figure}

Large language models (LLMs) have achieved remarkable success over the past two years~\cite{qwen3,deepseek,gpt4,gemini}, giving rise to a wide range of applications~\cite{personalagent,zhao2024comprehensive}. Among these, leveraging LLMs for building intelligent assistants~\cite{zhao2025reinforced,xi2025rise} has emerged as one of the most promising directions to support everyone's daily life and work.
Many efforts from both academia and industry have been devoted to this direction~\cite{agentsurvey,zhao2025exploring}, leading to early products such as Apple Intelligence\footnote{\url{https://www.apple.com/apple-intelligence/}} and the Meta AI app\footnote{\url{https://ai.meta.com/meta-ai/}}. 
A key factor in building such assistants is personalization. LLMs are originally designed under a one-size-fits-all paradigm; however, when integrated into individuals’ daily lives and work, they inevitably encounter diverse user preferences and needs~\cite{personalagent}. Personalizing LLMs—adapting them to accommodate individual differences—is essential for delivering high-quality service to enhance user experiences, giving rise to the emerging field of LLM personalization~\cite{personalizationsurvey1,personalizationsurvey2,personalizationsurvey3,personalizationsurvey4,personalizationsurvey5}.

For LLM personalization, activation steering has emerged as a promising and cost-effective solution~\cite{steerablechatbots,cos,cas,bipo}. This method directly modifies the LLM’s internal states to achieve desired outputs by adding a vector—a direction in the activation or hidden state space that can affect the characteristics of the model’s responses. Based on how the steering vector is computed for LLM personalization, two main methods have emerged. The first derives a user’s “style vector”~\cite{cas} by comparing historical user-authentic responses (containing both content and style) with 
user-agnostic responses (content-only) and then uses it to steer the model’s output toward the user’s style. The second computes an influence vector~\cite{cos} from the used historical user textual data on the model activations and amplifies it to strengthen personalization. Both approaches can effectively adjust model behavior to better align with user preferences, with promising results demonstrated across various scenarios.

Despite notable progress made by these efforts, we argue that a common limitation is that both of their steering vector computations rely blindly on all the historical user data, overlooking the fact that not all parts of the user data reflect the user’s preferences. Even though the textual data is produced by the user, each data piece often contains heterogeneous information. Different elements play different roles and can naturally show different degrees of relevance to user preferences~\cite{nextquill}. For example, there are some boilerplate parts~\cite{ziang}, e.g., common connecting words or sentences, which usually do not reflect user preference but only work as connecting functions. Treating all components as equally relevant risks dilutes the true preference signal and introduces noise into the steering vector, finally preventing the personalization quality.

In this work, we propose to disentangle preference-driven components from less preference-related ones, focusing on deriving the steering vector from the preference-driven components to enable more accurate personalization. This task is non-trivial, as there is no direct signal indicating which parts of the data are driven by user preferences. Inspired by prior studies~\cite{nextquill,cft}, we employ a causal analysis framework to address this challenge. As illustrated in Figure~\ref{fig:causal_graph}, the observed user data arises from the interaction between the user and the context. According to causal inference theory, determining whether a component is preference-driven requires measuring the extent to which it is influenced by the user’s inherent traits. This influence corresponds to the causal effect of the user on the observed data, quantified by the difference between the factual world and a counterfactual world where a reference user (without individual preference) generates the data. By leveraging the magnitude of these causal effects, we can effectively identify the preference-driven components of each user data instance.

To address this, we propose SteerX, a disentangled steering method inspired by causal analysis. SteerX follows a disentangling–smoothing–steering paradigm for LLM personalization: it disentangles preference-driven tokens through causal effect estimation, transforms them into smooth descriptions, and then leverages them in the steering process.
Specifically, we first estimate token-level causal effects: for each historical data instance, we represent the user trait using all other histories, then compare the factual prediction probabilities of the target history tokens (with the user trait) against their counterfactual predictions (without the user trait) to approximate causal effects via the LLM. Based on these effects, tokens are classified as either preference-driven or not.
Since the extracted preference-driven tokens are discrete and lack smooth semantics, we reconstruct them into coherent, fluent sentences to facilitate downstream steering. Finally, these smoothed preference-driven segments replace the original history data in existing steering methods to achieve personalization.

The main contributions of this work are summarized as follows:
\begin{itemize}[leftmargin=*]
\item To the best of our knowledge, we are the first to study the disentanglement of preference-driven and other components in user data to enable more accurate steering for LLM personalization.
\item We propose a novel disentangled steering method, \textbf{SteerX}, for LLM personalization, which leverages causal effects to separate preference-driven components from user data, followed by a smoothing–steering process to achieve personalization.
\item We apply SteerX to two existing representative steering methods, and extensive experiments on multiple real-world datasets demonstrate that it effectively enhances activation steering accuracy, thereby improving LLM personalization.
\end{itemize}

\section{Related Work}
\subsection{Personalized Text Generation}

With the rapid development of LLMs, these models have shown exceptional generative performance across a wide range of tasks, revolutionizing many areas of artificial intelligence~\cite{transformer,zhao2024pacar,chatbotarena,prism,synthesizeme,abbasiantaeb2024let}.
However, despite their impressive capabilities, LLMs still struggle to adapt to individual preferences and user needs.
As users increasingly demand personalized interactions, this limitation has sparked growing interest in LLM personalization~\cite{personalizationsurvey1,personalizationsurvey2,personalizationsurvey3,personalizationsurvey4,personalizationsurvey5}, with researchers focusing on how to tailor these models to provide more relevant, context-aware, and user-specific responses.

Recently, LLM personalization has been extensively explored across various domains, including image generation~\cite{pmg,pigeon,drc}, conversational agents~\cite{ldagent,THEANINE,TACITREE}, recommendation systems~\cite{automr,ye2025harnessing,igd}, and multimedia applications~\cite{uigan,music,puma}.
Among these, personalized text generation remains a fundamental and central topic within the field of LLM personalization~\cite{lamp,longlamp,dpl,dep}, as it directly impacts the quality and relevance of user interactions.
Existing methods for personalized text generation can be mainly categorized into two groups: 1) retrieval-based methods~\cite{hydra,cfrag}, where ROPG~\cite{ropg} proposes a retrieval selection model that dynamically chooses the best retriever to capture valuable personalized information, while DPL~\cite{dpl} focuses on extracting inter-user differences to improve LLM personalization; and 2) fine-tuning-based methods~\cite{personalizedpieces,proper}, where OPPU~\cite{oppu} learns a set of parameter-efficient adapters for each user, while PPlug~\cite{pplug} encodes the user history as a soft prompt.

\subsection{Causal Inference in LLMs}

Causal inference provides a principled framework to reason about cause-effect relationships, and its integration with LLMs has recently attracted increasing attention~\cite{thebookofwhy,zhu2024causal,causalsurvey}.
Unlike purely correlational approaches, causal methods aim to disentangle spurious correlations from genuine causal effects, which is crucial for a broader range of NLP tasks~\cite{kgd,cancausal,causalframework}.
For instance, in recommender systems, causal techniques have been applied to better model user-item interactions, mitigate biases, and improve personalization~\cite{pda,wu2023causality,cft}.
NextQuill~\cite{nextquill} is the first work to systematically apply causal inference to personalized text generation.
It proposes a unified causal framework that models user preference effects from both the model side and the data side, demonstrating the importance and effectiveness of causal inference for enhancing LLM personalization.

\subsection{Activation Steering}

Activation steering focuses on controlling the behavior of LLMs by directly injecting a direction vector into their internal activations during inference, rather than modifying model weights or input prompts~\cite{2022extracting,zou2023representation,alphasteer}.
This research is motivated by recent findings that LLMs can control specific semantics or behaviors via linear directions in their activation space~\cite{2024linear}.
Building upon this, several methods have been proposed: ActAdd~\cite{actadd} performs activation addition using contrastive-derived steering vectors for controlling sentiment and topic, while CAA~\cite{CAA} enhances steering precision by leveraging mass-mean activation differences.

Recently, some works~\cite{bipo,steerablechatbots} have begun exploring activation steering for LLM personalization, focusing on two main implementations: influence vector~\cite{cos} and style vector~\cite{cas}, which show promising potential for adapting model behavior to individual users.
However, these steering-based methods overlook the fact that user data is not entirely produced by the user and that each data piece often contains heterogeneous information, leading to suboptimal performance in LLM personalization.
To the best of our ability, our proposed SteerX addresses this limitation.
\section{Preliminary: Activation Steering}

\begin{figure}[t]
    \centering
    \includegraphics[width=\linewidth]{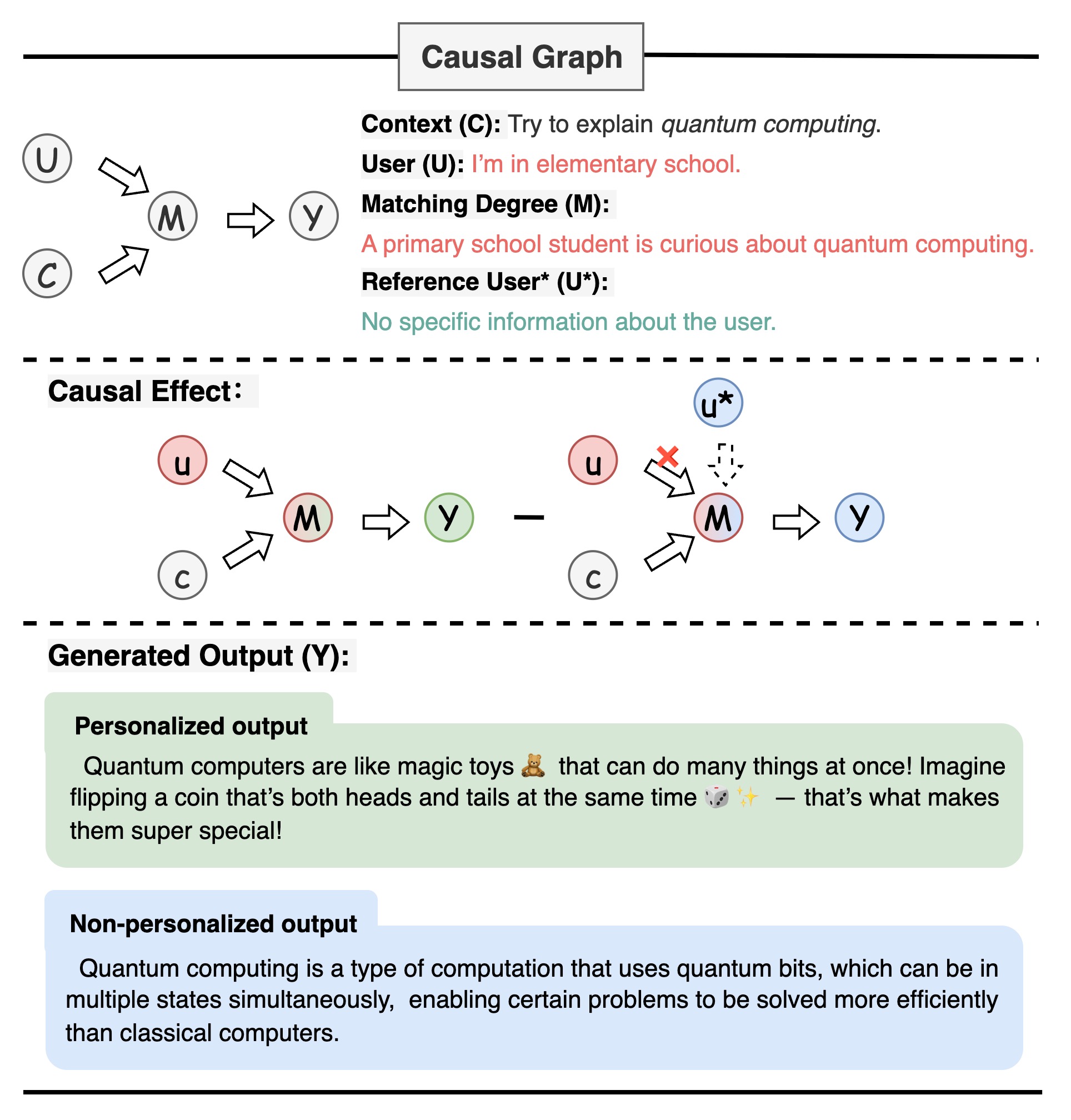}
    \caption{Illustration of our causal inference framework for personalized text generation.}
    \label{fig:causal_graph}
    \vspace{-1.2em}
\end{figure}
 
Let \(x \in \mathcal{X}\) denote the input (e.g., a user-specific query) and \(y \in \mathcal{Y}\) denote the target output.  
 \(P\) is the \textbf{task prompt} describing the desired task. 
Each historical interaction consists of an input \(x_i\) and its corresponding \emph{authentic} user-specific response \(h_i\).  
The user's total history is
\begin{equation}
H = \{(x_1, h_1), (x_2, h_2), \dots, (x_n, h_n)\}
\label{eq:user_history}
\end{equation}

\(\mathrm{\mathcal{LLM}_{logits}}(y \mid \cdot)\) denotes the output logits of the language model for \(y\) given the specified input condition.
\(\mathrm{\mathcal{LLM}_{hidden}}(\cdot)\) is the hidden state representation of the LLM at a chosen layer \(\ell\).

Activation steering modifies intermediate hidden states during inference to incorporate user-specific preferences.  
Current methods can be divided into two categories: \emph{influence vector} and \emph{style vector}.  
Both compute a steering direction \(a \in \mathbb{R}^d\) in the activation space and add it to the model’s hidden states during inference, but differ in how \(a\) is derived.

\subsection{Influence Vector}

\begin{figure*}[!t]
    \centering
    \includegraphics[width=1.0 \linewidth]{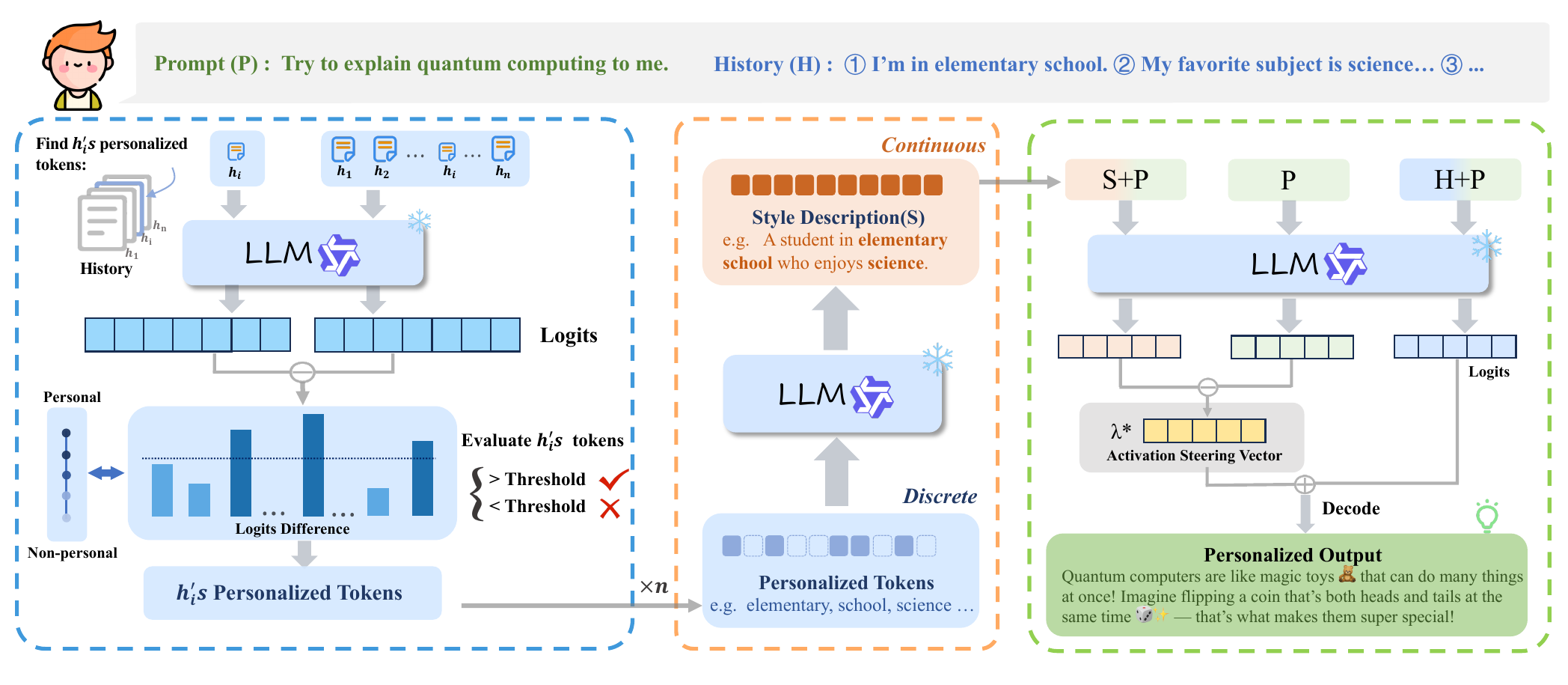}
    \vspace{-7mm}
    \caption{Overview of the proposed \methodnameshort framework for personalized text generation.  (\textbf{Left}) \emph{Personalized anchor token identification}: For each history \(h_i\), we apply a leave-one-history-out strategy and mark tokens with causal influence scores exceeding a predefined threshold as \emph{personalized tokens}.
    (\textbf{Middle}) \emph{Style profile generation}: All personalized tokens are transformed from a discrete token set into a continuous, coherent \emph{preference description} \(S\).  (\textbf{Right}) \emph{Personalization steering}: \(S\) is injected into the generation process. A personalization scaling factor \(\lambda\) controls the strength of steering based on the style vector.}
    \label{fig:main_framework}
    \vspace{-3mm}
\end{figure*}

The \textbf{influence vector} (IV) measures the change in the model's logits caused by the user's histories, a training-free steering mechanism.  
The user-specific preferences can be obtained by the logits difference between the output of LLM with the user histories $H$ (\textit{i.e.,} inputting both $H$ and prompt $P$) and the output of LLM without the user history (\textit{i.e.,} inputting $P$ only).
A representative activation steering method of this approach is Context Steering (CoS)~\cite{cos}, which applies influence vectors to control personalization strength directly from user histories.

Formally, let $\mathrm{\mathcal{LLM}_{logits}}(x, H ; P)$ and $\mathrm{\mathcal{LLM}_{logits}}(x, \varnothing ; P)$ denote the generation logits given the input $x$ with and without context, the activation steering direction is:
\begin{equation}
a_{IV} = \mathrm{\mathcal{LLM}_{logits}}(x, H ; P) - \mathrm{\mathcal{LLM}_{logits}}(x, \varnothing ; P).
\end{equation}
Scaling this direction by $\lambda$ can modulate the personalization strength during LLM generation:
\begin{equation}
\mathrm{Steer}_{IV}(x, H ; P) = \mathrm{\mathcal{LLM}_{logits}}(x, H ; P) + \lambda \cdot a_{IV}.
\end{equation}

\subsection{Style Vector}

The \textbf{style vector} (SV) encodes user-specific preferences derived from the user's historical responses, also as a training-free steering mechanism. 
It is derived by contrasting \emph{user-authentic} responses against \emph{user-agnostic} responses generated by a general-purpose LLM, isolating the stylistic component independent of task content.
A representative method is the  Contrastive Activation Steering (CAS)~\cite{cas} method, which gets the style direction of each instance and then aggregates them to obtain an overall style representation for the user.

Given each history pair $(x_i, h_i)$, a style-neutral response $\bar{h}_i$ is first generated with task prompt $P$:
\begin{equation}
\bar{h}_i = \mathcal{LLM}(x_i; P)
\label{eq:user_history}
\end{equation}

The style direction is computed as the mean hidden-state difference:
    \begin{equation}
    a_{SV} = \frac{1}{n} \sum_{i=1}^{n} \left[ \mathrm{\mathcal{LLM}_{hidden}}(h_i) - \mathrm{\mathcal{LLM}_{hidden}}(\bar{h}_i)\right].
    \end{equation}

During inference, steering is applied to obtain the personalized generation:
\begin{equation}
Steer_{SV}(x, H ; P) = \mathrm{\mathcal{LLM}_{hidden}}(x, H ; P) + \gamma \cdot a_{SV},
\end{equation}
where $\gamma$ controls personalized strength.

\noindent\textbf{\textit{Comparison.}}
As illustrated in Figure~\ref{fig:comparison}, the IV is computed by contrasting model outputs with and without the entire historical context for a given $(H, P)$ pair, capturing the global effect of history.
In contrast, the SV is computed at the instance level by contrasting  $(h_i, \bar{h}_i)$ pairs, then averaged across all histories to form a global style representation.

\section{Methodology}

To overcome the indiscriminate use of full user histories in existing activation steering methods, we propose \textbf{SteerX}, a three-stage (\emph{"disentangling–smoothing–steering"}) disentangled steering framework for enhancing personalized text generation.
As shown in~Figure~\ref{fig:main_framework}, 
grounded in causal inference theory, 
SteerX estimates token-level causal effects to identify preference-driven tokens (\emph{disentangling}), transforms these discrete signals into a coherent natural-language description (\emph{smoothing}), and leverages this representation to steer personalized LLM generation (\emph{steering}). Next, we first present the causal foundation of the work, followed by the detailed description of the three steps of SteerX.

\subsection{Task Formulation as Causal Graph}


Personalized text generation involves multiple intertwined factors, including the specific task context faced by the user, the user's inherent preferences, and the interaction between the two.
Therefore, directly quantifying personalized components from observed behavior is challenging, as it is entangled by other non-preference-related influences.
To explicitly disentangle the preference-driven components from these contextual effects, we formalize the process using a causal graph.

\noindent\textbf{\textit{Formulation.}}
We define the following variables:
\begin{itemize}[leftmargin=*]
    \item \textbf{Context (C): } 
    the task scenario and objectives (or items) that the user need to response.
    \item \textbf{User (U): } the underlying preferences or interests of user.
    \item \textbf{Matching (M): } the interaction (matching) between the current context \(C\) and the user \(U\).
    \item \textbf{Generated Output (Y): } the content produced by the user in response to the given context.
\end{itemize}

\noindent\textbf{\textit{Factual Process.}} 
As shown in Figure~\ref{fig:causal_graph}, the factual generation process can be expressed as:
\begin{equation}
H \rightarrow M \leftarrow C, \quad M \rightarrow Y
\label{eq:causal_graph}
\end{equation}
Here, \(U\) and \(C\) jointly determine the matching variable \(M\), which mediates the personalization effect on the generated output \(Y\).  

\noindent\textbf{\textit{Counterfactual Process.}}
To isolate the contribution of personalization, 
we need to define a counterfactual by intervening $U$ to a reference user \(u^{*}\), as shown in Figure~\ref{fig:causal_graph}: 
\begin{equation}
do(U = u^{*})
\label{eq:do_operator}
\end{equation}
Here, \(u^{*}\) corresponds to a generic user without individual preference.
This intervention severs the causal link from the original \(U=u\) to \(M\), while keeping \(C\) fixed.  
By comparing the output \(Y\) under the factual and counterfactual settings, we can obtain the personalization effect:
\begin{equation}
\begin{split}
    \Delta(Y) & = P(Y \mid do(U=u), C) - P(Y \mid do(U=u^{*}), C) \\
   & = P(Y \mid U=u, C) - P(Y \mid U=u^{*}, C)
\end{split}
\label{eq:causal_effect}
\end{equation}
The causal effect \(\Delta(Y)\) can be attributed to the personalization pathway \(U \rightarrow M \rightarrow Y\).

\subsection{Causal Analysis}
\label{sec:causal_effect_analysis}

In the context of LLM personalization, we approximate the \emph{user} \(U\) by the \emph{user history} \(H\), while the \emph{context} \(C\) corresponds to the task instructions and any accompanying situational information (including the item/objective the user needs to response to).  
Our goal is to isolate the \emph{preference-driven components} within a user’s data instance—i.e., the specific parts of the content that are causally attributable to the personalization pathway \(H \rightarrow M \rightarrow Y\).

To achieve this, we require a quantitative measure of personalization influence at the \emph{token level}.  
By comparing the generation behavior under factual and counterfactual conditions, we can determine how strongly each token reflects the user’s preference, enabling the identification of tokens that carry user-preference-related signals.

\vspace{+5pt}
\noindent\textbf{\textit{Token-Level Causal Effect.}}
After representing each user with their history, the token-level causal effect for $t$-th token $Y^t$ in $Y$ can be defined as follows, based on Equation~\eqref{eq:causal_effect}:
\begin{equation}
\Delta_t = \log p(Y^t \mid H, C) - \log p(Y^t \mid H^{*}, C)
\label{eq:delta}
\end{equation}
where $H^{*}$ denotes the corresponding reference that can be set to $\varnothing$ to indicate no preference.
A larger \(\Delta_t\) indicates a stronger influence from the personalization pathway \(H \rightarrow M \rightarrow Y\), which is obvious.



\vspace{+5pt}
\noindent\textbf{\textit{Anchor Preference Token Identification.}}
We can then define the set of \emph{anchor preference tokens} as:
\begin{equation}
\text{AnchorTokens}(y) = \{\, Y^t \mid \Delta_t \ge \tau \,\}
\label{eq:anchor_tokens}
\end{equation}
where \(\tau\) is a fixed threshold or is chosen adaptively so that the top-$k\%$ tokens with the highest \(\Delta_t\) values are selected, treating them as preference-driven tokens.

\subsection{Disentangling: Personalized Anchor Token Identification}\label{sec:disentangling}

Building on the causal effect analysis in Section~\ref{sec:causal_effect_analysis}, which identifies \emph{preference-driven tokens} in the generated output, we now turn to the \emph{input side} to locate the corresponding preference-driven components within the user history.  
The goal is to extract the parts of the history that contribute most to the personalization signal, thereby enabling the steering vector to be derived from truly preference-relevant content rather than from the entire history. 

\vspace{+5pt}


\noindent\textbf{\textit{Leave-One-History-Out Strategy.}}
Let $H = \{ h_1, \dots, h_n \}$ denote a user's history, omitting the input part $x_i$ of each data point here for brevity. 
To identify preference tokens for a given history $h \in H$, we treat $h$ as the outcome variable $Y$ in Equation~\eqref{eq:delta}, and represent the user with the remaining histories $H^{-} = H \setminus \{ h \}$. The causal effect of a token $h^t$ in $h$ is then computed as, based on Equation~\eqref{eq:delta}:
\begin{equation}
\Delta_t = \log p(h^t \mid H^{-}, C) - \log p(h^t \mid \varnothing, C),
\label{eq:delta_definition}
\end{equation}
where $p(\cdot)$ is estimated using LLMs.

Tokens with larger $\Delta_t$ are considered more preference-driven. Specifically, tokens whose attribution scores exceed a predefined threshold $\tau$ are collected to form the \emph{anchor preference token set} for history $h_i$:
\begin{equation}
\mathcal{T}_{\text{anchor}} = \{\, h^t \in h \mid \Delta_t \ge \tau \,\}
\label{eq:anchor_tokens}
\end{equation}
This procedure disentangles the preference-driven components of the user history from those less related to preferences.

\subsection{Smoothing: Style Description Generation}\label{sec:smoothing}

In Section~\ref{sec:disentangling}, we disentangled the preference-driven parts of the user history from the less preference-related ones.  
Although these tokens offer fine-grained attribution, they are discrete and fragmented, lacking global coherence and contextual continuity necessary for effective personalization.  

To steer LLM personalization more effectively, we transform the identified personalized anchor tokens into a compact, interpretable, and coherent representation.
Specifically, we apply a coherence transformation to \(\mathcal{T}_{\text{anchor}}\) to obtain the more personalized style profile:
\begin{equation}
s = \mathrm{Coherence}(\mathcal{T}_{\text{anchor}})
\label{eq:coherence_score}
\end{equation}
where \(\mathrm{Coherence}(\cdot)\) employs an LLM to generate a fluent, context-aware description that summarizes the key content/style reflected in the anchor tokens. 
Applying this process to each historical data point's anchor tokens yields the complete preference-driven portion of the all user history for steering, denoted by $S = \{ s_1, \dots, s_n \}$, where $s_i$ denotes the result for the $i$-th history.

\subsection{Steering: Personalized Generation}\label{sec:steering}

Following the \emph{disentangling–smoothing–steering} paradigm, the previous two stages isolated preference-driven components from the user history (\emph{disentangling}) and transformed them into a compact, coherent style description \(S\) (\emph{smoothing}).  
In this final \emph{steering} stage, we build on the activation steering paradigms—\emph{influence vector} and \emph{style vector}—to incorporate \(S\) as an explicit and interpretable personalization signal for guiding LLM generation at inference time.  
Unlike prior methods~\cite{cas,cos} that directly use raw user history or unstructured style representations, we replace the original history with the preference-driven parts \(S\), deriving the steering vector from preference-driven components to achieve more accurate personalization.


\vspace{+5pt}
\noindent\textbf{\textit{Inference Vector with Style Description.}}
For the influence vector setting, given a prompt \(P\) and treating \(S\) as the context, the steering direction is
\begin{equation}
F_{S,P}(x_i) = \mathcal{LLM}_{\mathrm{logits}}(x_i,  S; P) - \mathcal{LLM}_{\mathrm{logits}}(x_i, \varnothing; P)
\label{eq:inference_vector}
\end{equation}
and the steered logits become
\begin{equation}
\mathrm{Steer}_{IV}(x, H ; P) = \mathcal{LLM}_{\mathrm{logits}}(x_i, H; P) + \lambda \cdot F_{S,P}(x_i)
\label{eq:cos_formula}
\end{equation}
where \(\lambda\) controls the strength of \(S\)’s influence.

\noindent\textbf{\textit{Style Vector with Style Description.}}
For the style vector setting, we enhance CAS~\cite{cas} by incorporating the style description \(S\) into the hidden-state difference computation. 
Given each history pair \((x_i, h_i)\), we first generate a style-neutral response \(\bar{h}_i = \mathrm{\mathcal{LLM}}(x_i; P)\). 
The refined style direction is then computed by averaging three hidden-state differences: the user-authentic response vs. style-neutral, the style description vs. style-neutral, and (optionally) the original CAS term. 
Formally, at layer \(\ell\):
\begin{equation}
\begin{aligned}
a_{SV} &= \frac{1}{n} \bigg[
\sum_{i=1}^{n} \gamma_1 \cdot\left( \mathrm{\mathcal{LLM}_{hidden}}(h_i) - \mathrm{\mathcal{LLM}_{hidden}}(\bar{h}_i) \right) \\
&\quad + (1-\gamma_1)\cdot\left( \mathrm{\mathcal{LLM}_{hidden}}(s_i) - \mathrm{\mathcal{LLM}_{hidden}}(\bar{h}_i) \right)
\bigg]
\end{aligned}
\end{equation}

During inference, steering is applied to the hidden states as:
\begin{equation}
\mathrm{Steer}_{SV}(x, H ; P) = \mathrm{\mathcal{LLM}_{hidden}}(x, H; P) + \gamma_2 \cdot a_{SV}
\end{equation}
where \(\gamma_1, \gamma_2\) controls the steering strength. 
Here, \(S\) can also be used to filter or select history entries most aligned with the target style, further improving the relevance and precision of the steering signal.

By introducing the structured style profile \(S\) into activation steering, we enable fine-grained personalization control without retraining the base model, directly linking more preference-driven parts of history content to output personalization, 
and condensing personalization cues into a single steering representation for flexible control during generation.

\section{Experiments}

\begin{table*}[!h]
\centering
\caption{Performance comparison between baseline methods and the proposed SteerX across two personalized text generation tasks using LLMs of two different sizes. For each LLM, the best result is shown in \textbf{bold}. Higher values indicate better performance across all metrics. \textit{Impr.} denotes the relative average improvement across four metrics for each method compared with the Non-Person baseline.}
\label{main_tab}
\renewcommand{\arraystretch}{0.95}
\resizebox{\textwidth}{!}{%
\begin{tabular}{@{}c*{11}{c}@{}}
\toprule
\multicolumn{2}{c}{\textbf{Datasets ($\rightarrow$)}} &
\multicolumn{5}{c}{\textbf{Review Generation}} &
\multicolumn{5}{c}{\textbf{Topic Writing}} \\
\cmidrule(lr){3-7} \cmidrule(lr){8-12}
\multicolumn{2}{c}{\textbf{Methods ($\downarrow$)}} &
ROUGE-1 & METEOR & BLEU & BERTScore & \textit{Impr.} $\uparrow$ &
ROUGE-1 & METEOR & BLEU & BERTScore & \textit{Impr.} $\uparrow$ \\

\midrule

\multicolumn{2}{l}{\texttt{Qwen3-8B} (Non-Perso)} & 0.2903 & 0.1911 & 1.9381 & 0.4646 & \textbf{---} & 0.2380 & 0.1649 & 1.1104 & 0.4448 & \textbf{---} \\
\multicolumn{2}{l}{\hspace{0.1em} + RAG} & 0.3061 & 0.2090 & 2.8455 & 0.4810 & 16.29\% & 0.2527 & 0.1913 & 2.5582 & 0.4587 & 38.92\% \\
\midrule
\multicolumn{2}{l}{\hspace{0.1em} + CAS} & 0.3119 & 0.2099 & 2.8429 & 0.4880 & 17.25\% & 0.2554 & 0.1896 & 2.6874 & 0.4581 & 41.83\% \\
\rowcolor[HTML]{fcf1e2} 
\multicolumn{2}{l}{\hspace{0.1em} + SteerX (CAS)} & 0.3118 & 0.2140 & 3.1307 & \textbf{0.4898} & 21.59\% & 0.2624 & 0.1940 & 2.6344 & 0.4600 & 42.14\% \\
\midrule
\multicolumn{2}{l}{\hspace{0.1em} + COS} & 0.3149 & 0.2168 & 3.1092 & 0.4839 & 21.63\% & 0.2625 & 0.2086 & 3.0324 & 0.4604 & 53.35\% \\
\rowcolor[HTML]{fcf1e2} 
\multicolumn{2}{l}{\hspace{0.1em} + SteerX (COS)} & \textbf{0.3156} & \textbf{0.2238} & \textbf{3.5912} & 0.4848 & \textbf{28.87\%} & \textbf{0.2658} & \textbf{0.2089} & \textbf{3.2097} & \textbf{0.4617} & \textbf{57.81\%} \\

\midrule
\midrule

\multicolumn{2}{l}{\texttt{Qwen3-14B} (Non-Perso)} & 0.2835 & 0.1884 & 1.8485 & 0.4641 & \textbf{---} & 0.2648 & 0.1952 & 1.8238 & 0.4509 & \textbf{---} \\
\multicolumn{2}{l}{\hspace{0.1em} + RAG} & 0.3143 & 0.2184 & 2.8731 & 0.4845 & 21.65\% & 0.2893 & 0.2221 & 2.4741 & 0.4701 & 15.74\% \\
\midrule
\multicolumn{2}{l}{\hspace{0.1em} + CAS} & 0.3205 & 0.2199 & 2.9782 & 0.4847 & 23.83\% & 0.2918 & 0.2148 & 2.5982 & 0.4711 & 15.48\% \\
\rowcolor[HTML]{fcf1e2} 
\multicolumn{2}{l}{\hspace{0.1em} + SteerX (CAS)} & \textbf{0.3209} & 0.2191 & 3.0231 & \textbf{0.4861} & 24.44\% & 0.2937 & 0.2154 & 2.5961 & 0.4728 & 17.12\% \\
\midrule
\multicolumn{2}{l}{\hspace{0.1em} + COS} & 0.3146 & 0.2184 & 3.1719 & 0.4838 & 25.68\% & 0.2861 & 0.2237 & 2.0095 & 0.4674 & 9.12\% \\
\rowcolor[HTML]{fcf1e2} 
\multicolumn{2}{l}{\hspace{0.1em} + SteerX (COS)} & 0.3201 & \textbf{0.2243} & \textbf{3.5787} & 0.4855 & \textbf{32.54\%} & \textbf{0.3063} & \textbf{0.2329} & \textbf{3.6407} & \textbf{0.4749} & \textbf{34.98\%} \\

\bottomrule
\end{tabular}%
}
\vspace{-0.5em}
\end{table*}

In this section, we conduct extensive experiments in real-world datasets to answer the following research questions:

\begin{itemize}[leftmargin=*]
    \item \textbf{RQ1:} How does our proposed SteerX perform on the personalized text generation tasks compared to existing baseline methods?
    \item \textbf{RQ2:} What is the contribution of each individual design of SteerX to its overall performance?
    \item \textbf{RQ3:} 
    How do the specific hyperparameters and settings in SteerX affect its overall performance?
    \item \textbf{RQ4:} 
    In what ways does SteerX illustrate its personalized generation capabilities through qualitative case studies?
\end{itemize}

\subsection{Experimental Setup}

\noindent
\textbf{Datasets.} 
Building upon prior work, our experiments focus on the review generation\footnote{Processed dataset: \url{https://huggingface.co/datasets/SnowCharmQ/DPL-main}~\cite{dpl}.}~\cite{amazondataset} and topic writing\footnote{Processed dataset: \url{https://huggingface.co/datasets/LongLaMP/LongLaMP}~\cite{longlamp}.}~\cite{tldr}, which serve as two representative personalized text generation tasks.
Following the standard settings of OPPU~\cite{oppu}, we form the test dataset by selecting the 100 most active users with the longest historical records.
For review generation, we use the \textit{Movies \& TV} category from Amazon, and for topic writing, we use \textit{Reddit} posts.


\vspace{0.4em}
\noindent
\textbf{Baselines.} We conduct a comprehensive comparison between SteerX and a set of existing baseline methods:
\begin{itemize}[leftmargin=*]
    \item \textbf{Non-Perso:}
    This represents the non-personalized setting, where all user-specific information is excluded from the model input.
    \item \textbf{RAG~\cite{lamp}:} 
    This method employs a retrieval approach to extract key user-specific information, which is then used as instructional context for personalization.
    \item \textbf{COS (Context Steering)~\cite{cos}:} 
    This method amplifies the influence of context by comparing LLM outputs with and without the context and linearly scales this effect to control the degree of personalization at inference time.
    \item \textbf{CAS (Contrastive Activation Steering)~\cite{cas}:}
    This method generates style-agnostic responses for a user’s history with a base LLM, derives style vectors by contrasting hidden activations between real and neutral responses, and steers generation at inference via linear activation interventions using these vectors.
\end{itemize}
Notably, COS and CAS are the most recent state-of-the-art steering methods, representing the influence vector and style vector approaches, respectively. We apply our method to both methods and compare them with their original versions to verify the superiority of our disentangled steering over different LLM backbones.

\vspace{0.4em}
\noindent
\textbf{Evaluation metrics.} 
To provide a rigorous and comprehensive evaluation of our method, we compute multiple metrics that quantify the correspondence between the generated outputs and the ground-truth data.
Specifically, we use standard metrics commonly adopted in personalized text generation tasks~\cite{lamp,longlamp,reviewllm}, including ROUGE-1~\cite{rouge}, METEOR~\cite{meteor}, BLEU\footnote{We utilize the standard \texttt{SacreBLEU}~\cite{sacrebleu} library to calculate the BLEU score here: \url{https://github.com/mjpost/sacrebleu}.}~\cite{bleu}, and BERTScore\footnote{We adopt the \texttt{led-base-16384}~\cite{longformer} model to obtain embeddings for calculating the BERTScore: \url{https://huggingface.co/allenai/led-base-16384}.}~\cite{bertscore}.
Among these, we take METEOR as the primary evaluation metric owing to its extensive adoption in personalized text generation.


\vspace{0.4em}
\noindent
\textbf{Implementation Details.} 
We conduct experiments on two open source LLMs of different sizes~\cite{qwen3}: \texttt{Qwen3-8B}\footnote{\url{https://huggingface.co/Qwen/Qwen3-8B}} and \texttt{Qwen3-14B}\footnote{\url{https://huggingface.co/Qwen/Qwen3-14B}}.
We use the \texttt{Contriever}~\cite{contriever} model to perform a retrieval-based ranking of the current input prompt against the user's historical samples. 
The top two ranked samples are selected as historical context for all included methods. 
Subsequently, these two historical samples are combined with the current context and embedded using the \texttt{Qwen2.5-1.5B}\footnote{\url{https://huggingface.co/Qwen/Qwen2.5-1.5B}}~\cite{qwen25}. 
We then apply the method described in Section~\ref{sec:causal_effect_analysis} to obtain personalized anchor tokens. 
After generating anchor tokens from historical samples, we concatenate them into a single sequence and feed it into the \texttt{Qwen3-14B}~\cite{qwen3} model to produce a fluent, context-aware style description that captures the user’s stylistic and preference tendencies.
Finally, we generate the personalized output based on the method described in Section~\ref{sec:steering}.
As for the hyperparameters, for the review generation task, we set $(\gamma_1, \gamma_2, l, \lambda, \tau)$ to $(0.4,0.4,20,0.7,1.0)$ for the \texttt{Qwen3-8B} model and $(0.4,0.1,20,0.7,1.0)$ for the \texttt{Qwen3-14B} model, while for the topic writing task, we set $(\gamma_1, \gamma_2, l, \lambda, \tau)$ to $(0.1,0.1,32,0.75,0.7)$ for the \texttt{Qwen3-8B} model and $(0.1,0.1,20,0.3,0.7)$ for the \texttt{Qwen3-14B} model.
For inference, we disable sampling and always select the token with the highest probability, with the maximum generation length set to 2048 tokens.
Experiments are run on NVIDIA A100 and A800 GPUs with 80GB of GPU memory.

\subsection{Main Results (RQ1)}

We begin by evaluating the overall performance of all compared methods.
The main experimental results across the two personalized text generation tasks are presented in Table~\ref{main_tab}, from which we can draw the following observations:

\begin{itemize}[leftmargin=*]
    \item \textbf{Leveraging user history boosts the model’s ability to generate personalized text.}
    For both LLM sizes, Non-Person demonstrates the weakest personalization performance across two tasks, reflecting its inability to utilize user-specific contextual information to guide personalized generation.
    In contrast, RAG achieves notable gains by retrieving and incorporating user-specific data, producing more tailored and relevant output.
    \item \textbf{Activation steering approaches enhance personalization by directly modulating model behavior.}
    Methods including CAS and COS achieve general achievements in both personalized tasks compared to RAG.
    By injecting user-specific directional signals into the model's internal activations, these approaches enable the LLM to produce outputs that more accurately reflect individual user preferences, highlighting the promise of steering-based personalization.
    \item \textbf{SteerX achieves further improvements on LLMs of different sizes across tasks and metrics.} By disentangling the preference-driven component from user data for steering, our method enhances the quality of the steering vector, thereby improving personalization and generally outperforming the original steering method. Moreover, these improvements persist across diverse LLM backbones and steering methods, demonstrating both the effectiveness and the broad applicability of our approach.
\end{itemize}

\begin{figure}[h]
    \centering
    \includegraphics[width=\linewidth]{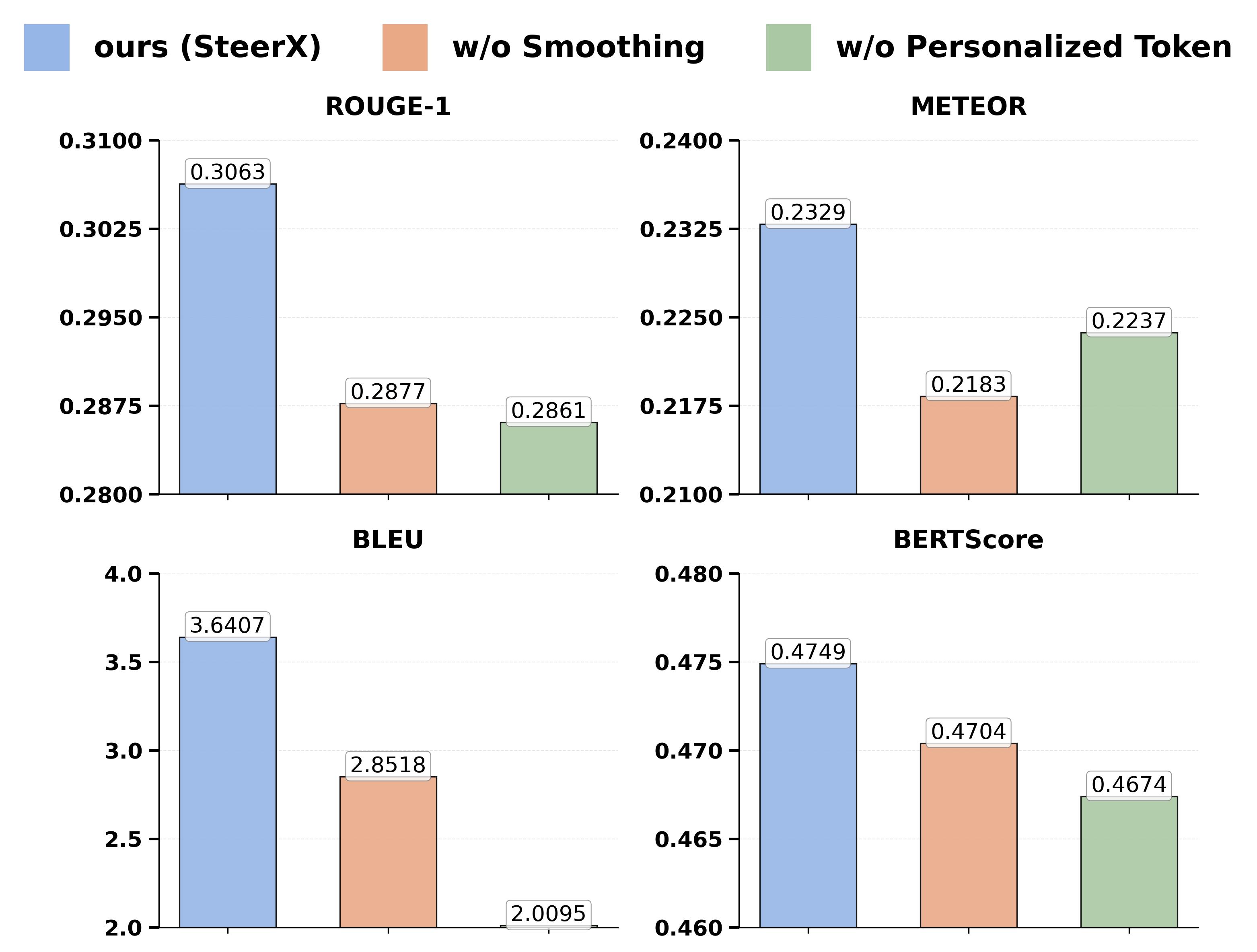}
    \caption{Ablation results on the topic writing task, comparing the full SteerX framework with two reduced variants: (1) w/o Smoothing and (2) w/o Personalized Token. Performance is reported across ROUGE-1, METEOR, BLEU, and BERTScore.}
    \label{fig:sum_vs_nosum}
\end{figure}
\subsection{Ablation Studies (RQ2)}
To assess the contribution of each component in our \emph{disentangling–smoothing–steering} paradigm, we compare three variants on topic writing dataset:  
(1) \textbf{w/o Personalized Token}: the strong personalized baseline without the token-level disentangling step;  
(2) \textbf{w/o Smoothing}: using only the disentangled personalized anchor tokens for steering, without coherence transformation;  
(3) \textbf{Ours (SteerX)}: our proposed model with preference smoothing via the preference description \(S\).  
The results are presented in Figure\ref{fig:sum_vs_nosum}.

\noindent\textbf{Effectiveness of Personalized Tokens.}  
Compared with \textbf{w/o Personalized Token}, the \textbf{w/o Smoothing} variant already yields notable improvements across all metrics, demonstrating that the identified anchor tokens indeed capture meaningful user preferences. This validates the effectiveness of our causal disentangling step in isolating preference-driven signals from noisy history.

\noindent\textbf{Benefit of Smoothing.}  
When replacing the discrete tokens with the smoothed preference description \(S\) in \textbf{Ours (SteerX)}, we observe further gains in all metrics. This indicates that the continuous, coherent representation better preserves contextual semantics and provides a stronger, more stable steering signal during generation. The improvement is consistent across LLM sizes, highlighting that the smoothing step not only enhances personalization quality but also improves robustness.

\subsection{In-Depth Analysis (RQ3)}
\subsubsection{Token-Level Causal Influence Analysis}

\begin{figure*}[t]
    \centering
    \includegraphics[width=0.96 \linewidth]{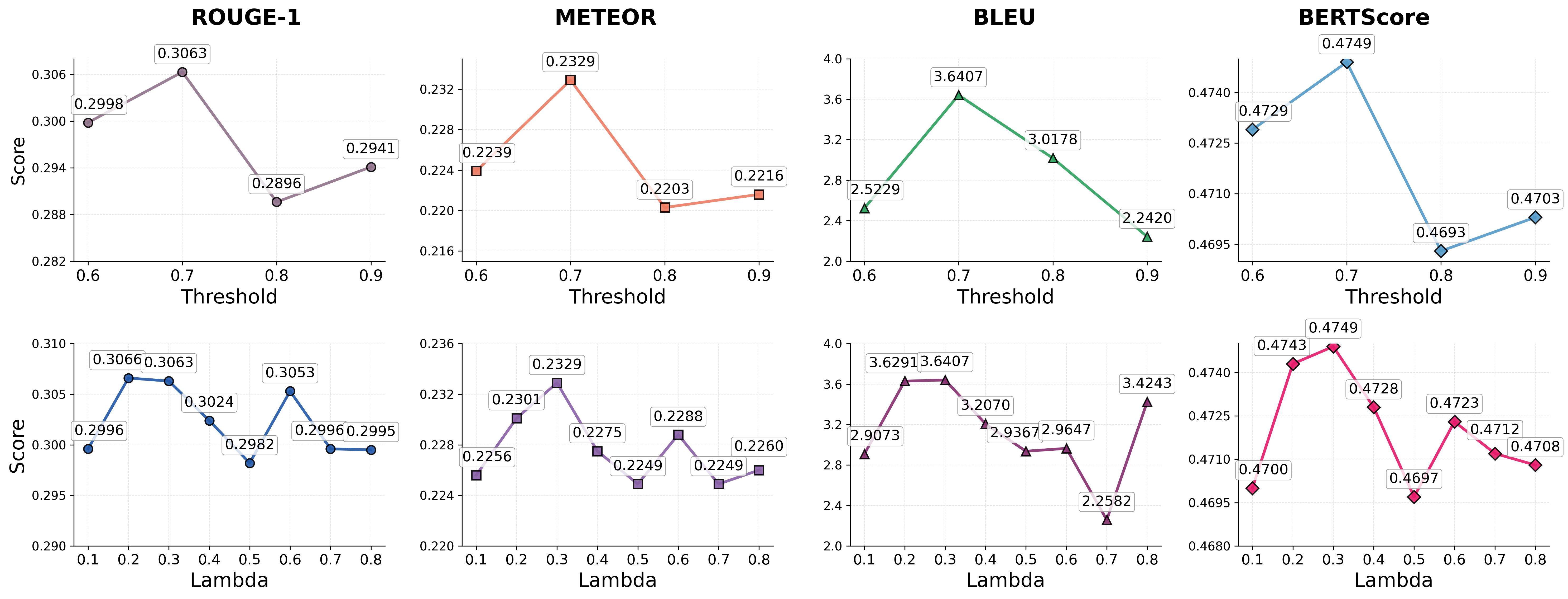}
    \caption{Effects of different hyperparameters on SteerX (COS) for the topic writing task: (1) upper panel — varying the causal effect threshold $\tau$; (2) lower panel — varying the steering coefficient $\lambda$. Performance is evaluated using four metrics: ROUGE-1, METEOR, BLEU, and BERTScore.}
    \label{fig:threshold_analysis}
    \vspace{-0.5em}
\end{figure*}

\begin{figure}[h]
    \centering
    \includegraphics[width=1.0 \linewidth]{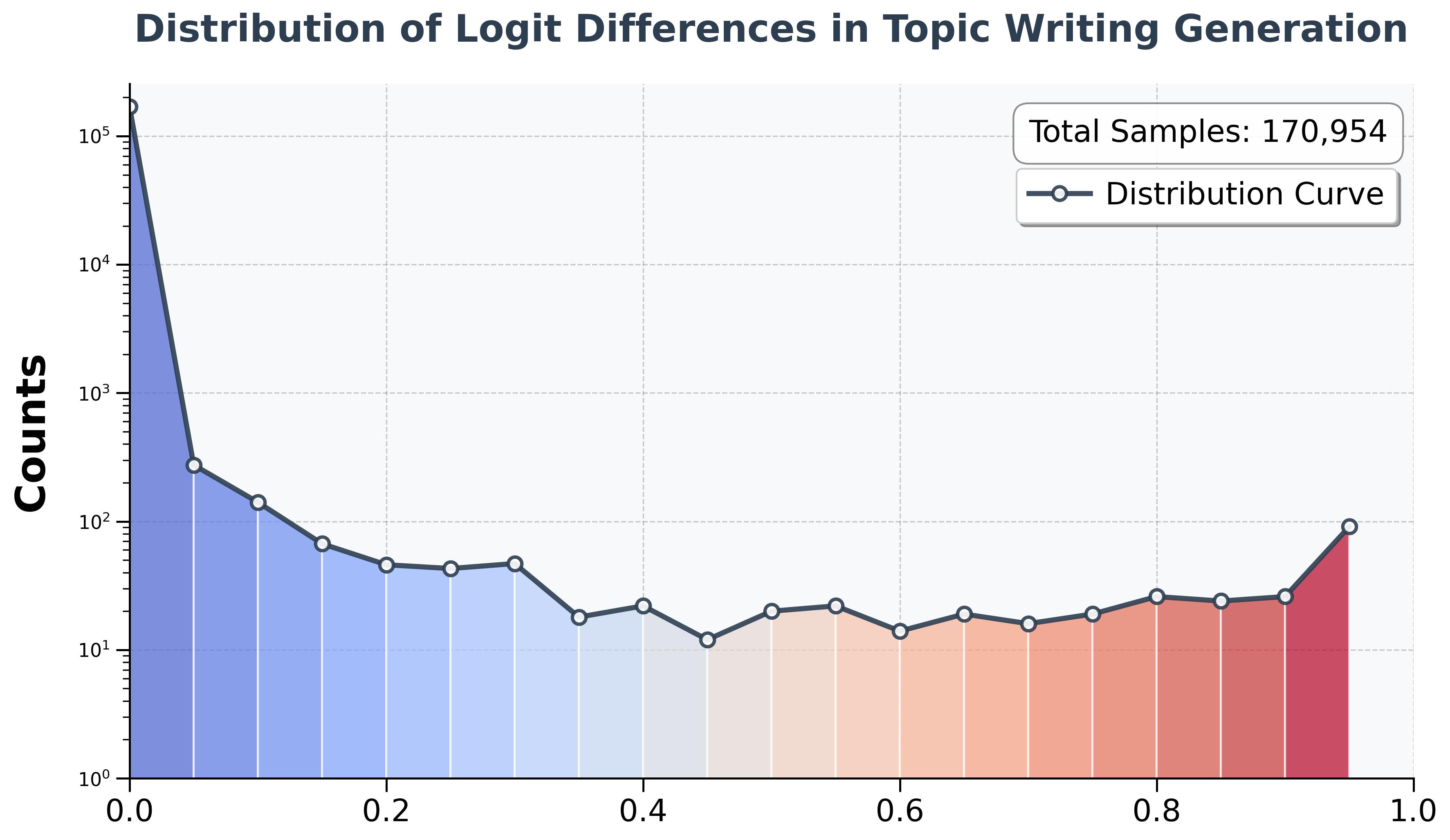}
    \caption{Distribution of token-level causal influence scores, quantified as logit differences between factual (with history) and counterfactual (without history) settings on the topic writing task. A subset of high scores corresponds to personalization-relevant tokens identified by SteerX.}
    \label{fig:token_diff_distribution}
    \vspace{-1.2em}
\end{figure}

To examine the behavior of our disentangling stage, we analyze the distribution of token-level causal effects, computed as the logit difference between the \emph{factual} setting (with full user history) and the \emph{counterfactual} setting (with the history removed).  
This difference, as defined in Section~\ref{sec:causal_effect_analysis}, quantifies how strongly each token is influenced by personalization.

Figure~\ref{fig:token_diff_distribution} visualizes the distribution of causal effect scores across all generated tokens.  
We observe that a large subset of tokens exhibits low causal influence, while a smaller subset forms a long-tailed distribution with significantly higher scores.  
These high-influence tokens correspond to the \emph{preference-driven components} identified by our method, which we term \emph{personalized anchor tokens}.  
This highlights the necessity of our causal analysis for isolating truly personalization-relevant content from noisy background information.

\subsubsection{Effect of Threshold \(\tau\) on Personalized Token Selection}


We further investigate the sensitivity of SteerX to the causal influence threshold \(\tau\), which determines which tokens are retained as personalized anchor tokens. The upper part of Figure~\ref{fig:threshold_analysis} reports performance (ROUGE-1, METEOR, BLEU, BERTScore) on the development set for \(\tau \in \{0.6, 0.7, 0.8, 0.9\}\).

Across all metrics, \(\tau=0.7\) consistently yields the best or near-best results, indicating a balanced trade-off between retaining sufficient personalization signals and excluding noisy tokens. Lower thresholds (e.g., \(\tau=0.6\)) include more tokens but also admit irrelevant content, diluting the personalization signal. Conversely, higher thresholds (\(\tau=0.8, 0.9\)) result in fewer selected tokens, improving precision but risking the omission of useful preference cues, which leads to performance drops in BLEU and ROUGE.

Overall, the performance curve remains relatively smooth, suggesting that SteerX is robust to moderate variations in \(\tau\). These findings highlight that a well-chosen threshold is essential for isolating preference-driven components while preserving sufficient coverage for coherent personalization.

\subsubsection{Effect of the Steering Strength \(\lambda\)}

We analyze the sensitivity of SteerX to the scaling factor \(\lambda\) that controls the magnitude of activation steering in the hidden-state space. The lower part of Figure~\ref{fig:threshold_analysis} reports ROUGE-1, METEOR, BLEU and BERTScore on the development set for \(\lambda \in \{0.1, 0.2, 0.3, 0.4, 0.5, 0.6, 0.7, 0.8\}\). Performance peaks in the light-to-moderate regime (\(\lambda \approx 0.2\!-\!0.3\); e.g., ROUGE-1 and BLEU reach their maxima near \(\lambda=0.2\!-\!0.3\)), then gradually declines as \(\lambda\) increases. This pattern indicates a trade-off: small \(\lambda\) under-utilizes the preference signal, whereas large \(\lambda\) over-amplifies it and perturbs content fidelity. The curves are relatively smooth, suggesting robustness to moderate shifts in \(\lambda\). Overall, these results support that steering in the activation space has a meaningful “sweet spot,” where the learned direction remains valid while the scaling preserves a good balance between personalization strength and task performance.

\subsubsection{Static LIWC Lexicon vs. Adaptive Causal Token Identification}

To further validate the effectiveness of our personalized anchor token identification, we compare SteerX with a LIWC-based approach. LIWC~\cite{pennebaker2015development} is an expert-crafted static lexicon widely used in psycholinguistics, containing predefined categories for personalization. These categories are designed by domain experts to map linguistic features to psychological traits and are applied uniformly across all text without adaptation to individual users. In our comparison, the LIWC-based method flags tokens in the generated text that match entries in its personalized dictionary, treating them as stylistic tokens.

\begin{figure}[!t]
    \centering
    \includegraphics[width=0.88 \linewidth]{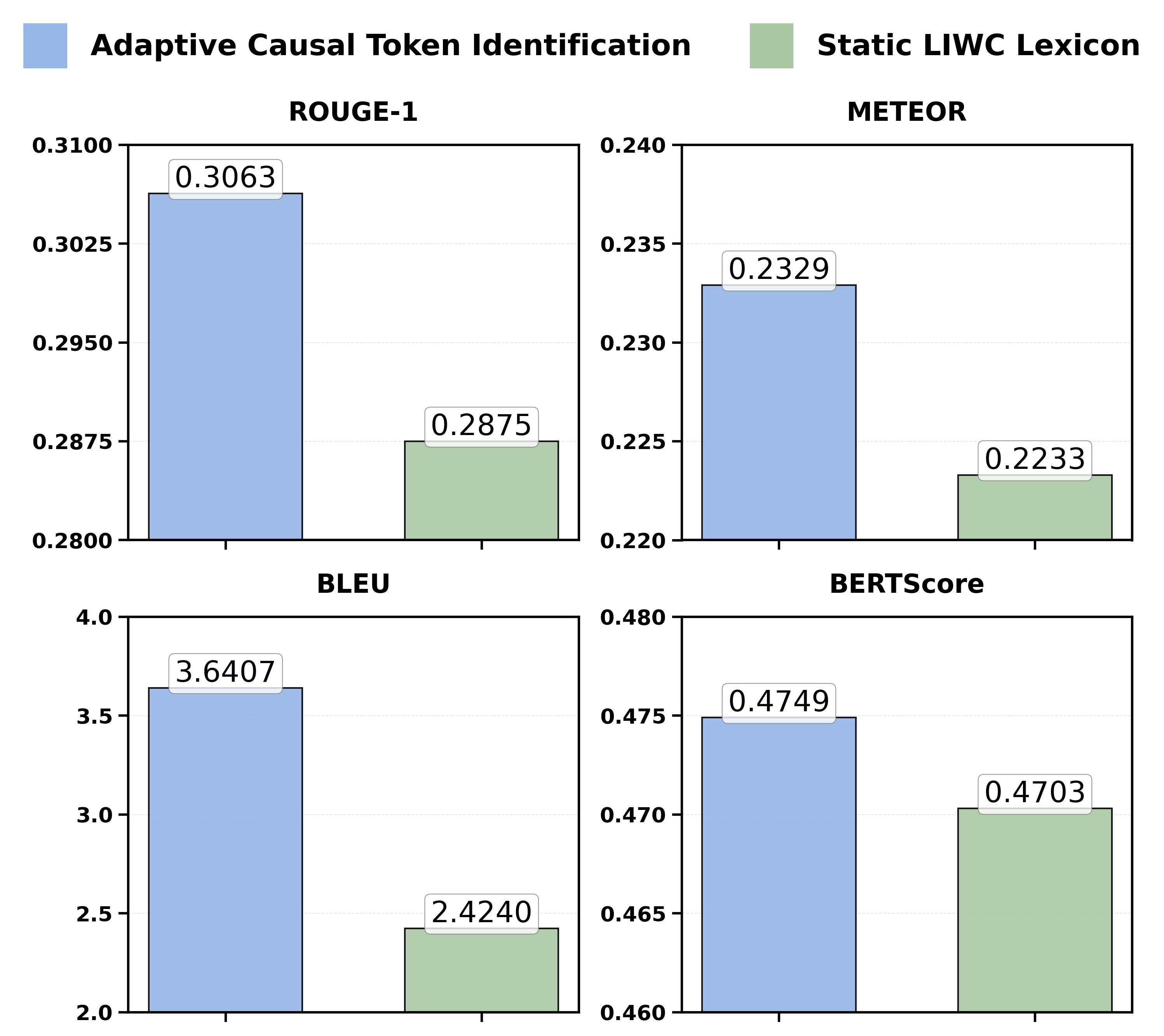}
    \caption{Performance comparison between our proposed adaptive causal token identification (\methodnameshort) and the static LIWC lexicon across four evaluation metrics: ROUGE-1, METEOR, BLEU, and BERTScore.}
    \label{fig:liwc}
    \vspace{-1em}
\end{figure}

Figure~\ref{fig:liwc} shows that SteerX consistently outperforms LIWC across all four metrics (ROUGE-1, METEOR, BLEU, and BERTScore) with absolute gains. This performance gap highlights two critical differences:
(1) LIWC’s static nature limits its ability to capture domain-specific or idiosyncratic user preferences that fall outside its fixed vocabulary.
(2) Our causal-effect-based identification dynamically adapts to each user’s history, detecting tokens with empirically validated influence on the model’s generation.

Besides, LIWC yields a sparse set of matched tokens, often missing subtle stylistic or topical cues. SteerX produces a richer and more contextually aligned token set, enabling finer-grained and more accurate personalization. This adaptivity is crucial in open-domain generation, where user style often diverges from standardized lexicons.
This analysis highlights the advantage of SteerX in learning personalized token representations directly from causal influence signals, rather than relying on fixed, expert-defined resources.

\subsection{Case Study (RQ4)}

To illustrate the effectiveness of SteerX in personalized review generation, we present a representative example from the review generation task. Following our \emph{disentangling–smoothing–steering} paradigm, SteerX first \textbf{disentangles} the user’s history by identifying \emph{personalized anchor tokens} through causal-effect analysis. These tokens are then \textbf{smoothed} into a coherent \emph{style description}, which serves as an explicit personalization signal. Finally, this description is \textbf{used to steer} the generation process at inference time.

\begin{figure}[!t]
    \centering
    \includegraphics[width=0.95 \linewidth]{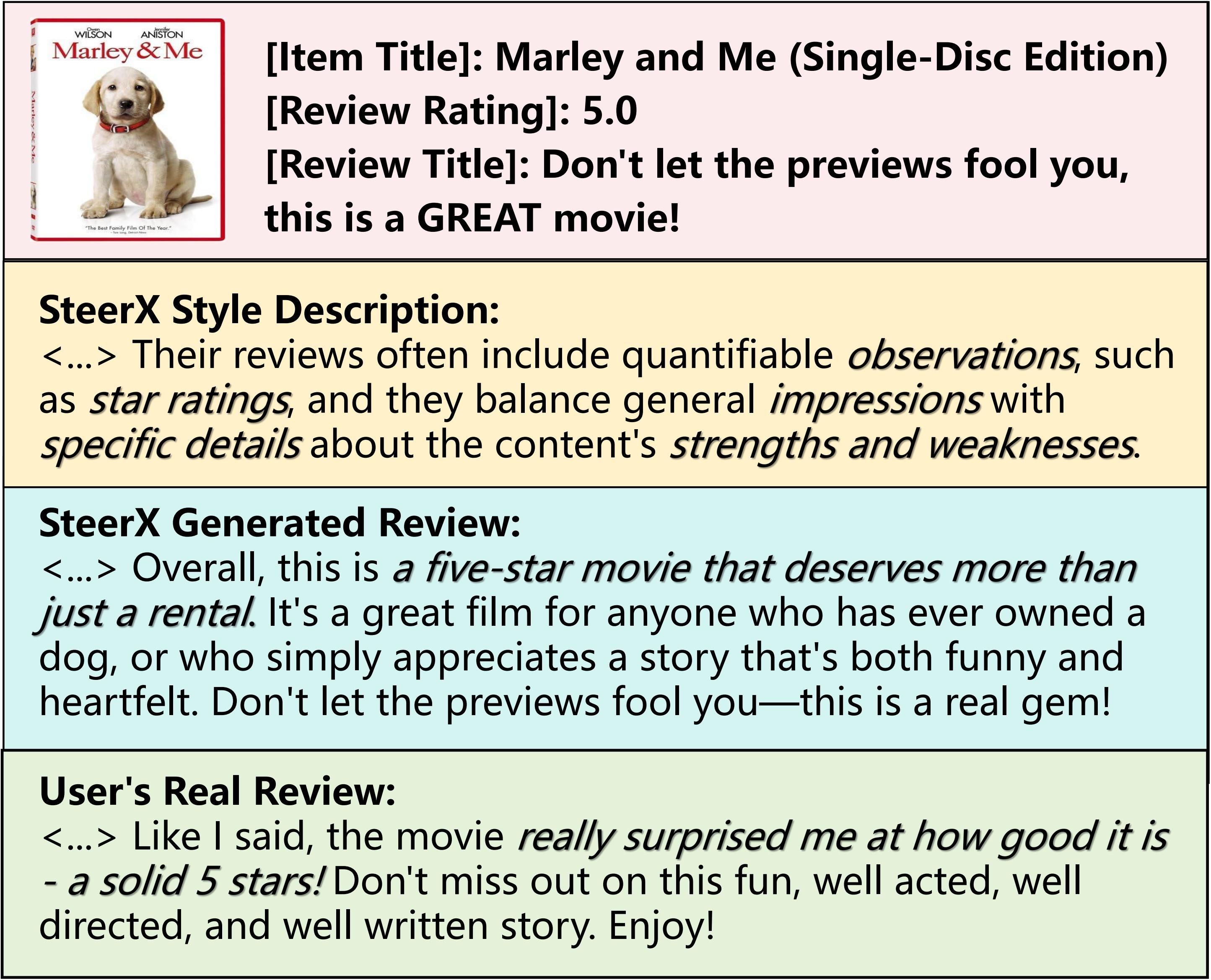}
    \caption{Case study of SteerX’s personalized review generation under the disentangling–smoothing–steering paradigm. SteerX preserves lexical overlap, sentiment intensity, and informative detail, aligning closely with the user’s preferences.}
    \label{fig:case_study}
    \vspace{-2em}
\end{figure}

In the example shown in Figure~\ref{fig:case_study}, the style description highlights the user’s tendency to blend emotional engagement with factual evaluation—combining expressive sentiment (e.g., “real gem”) with concrete assessments (e.g., “five-star movie”). The generated review by SteerX aligns closely with the user’s authentic review, not only reproducing the evaluative tone and stylistic markers but also capturing the rhythm and phrasing patterns characteristic of the user’s past writing. Besides, it preserves lexical overlap with the user’s expressions, integrates sentiment intensities, and provides informative detail. This case demonstrates SteerX’s ability to go beyond topical matching, generating text that faithfully reflects both the user’s preferences and their distinctive communicative style—enhancing authenticity, engagement, and user alignment in personalized content.

\section{Conclusion}


In this work, we introduced SteerX, a disentangled steering framework for LLM personalization that addresses key limitations of existing activation steering methods. Rather than constructing steering vectors from an entire user history—risking the inclusion of preference-irrelevant content—SteerX adopts a causal inference perspective to explicitly isolate \emph{preference-driven} components. By estimating token-level causal effects between factual and counterfactual histories, our approach pinpoints the tokens most influenced by intrinsic user traits. These high-impact tokens are then smoothed into coherent semantic descriptions, ensuring interpretability, reducing noise amplification, and enhancing compatibility with downstream steering.
Extensive evaluations on multiple real-world personalization datasets and two representative steering backbones confirm that SteerX delivers consistent improvements on ROUGE, METEOR, BLEU, and BERTScore. In-depth analyses highlight that the causal disentanglement stage effectively filters out irrelevant background signals, while the smoothing stage preserves fluency and semantic consistency—together enabling more precise and robust steering vector construction.

\vspace{0.8em}
\noindent\textbf{\textit{Future Potential.}} SteerX offers a generalizable paradigm for causally grounded personalization, where steering is informed by the true drivers of user-specific behavior rather than surface correlations. This opens new opportunities for integrating fine-grained causal analysis into large-scale LLM personalization pipelines, enabling adaptive, interpretable, and privacy-conscious personalization strategies. Future work may extend this framework to multimodal personalization and real-time interactive systems, further advancing the alignment of LLM outputs with diverse user needs and preferences.

\clearpage

\section*{Ethical Consideration}
Our method relies on the collection and analysis of users’ historical interaction data, which necessitates the application of strict security measures and adherence to robust ethical standards. The creation of user profiles for personalization purposes carries inherent risks, such as unauthorized surveillance or targeted manipulation. To mitigate these risks, the approach should integrate comprehensive safeguards, including secure data storage, effective anonymization, and clear mechanisms for obtaining informed user consent.
The handling of personal history data requires careful oversight to ensure it is not misused and remains protected throughout its lifecycle.
All experiments in this work utilize open-source datasets. The research upholds principles of data integrity, privacy protection, and ethical responsibility, ensuring the fair and transparent use of open-source resources.

\bibliographystyle{ACM-Reference-Format}
\bibliography{refs}

\end{document}